\documentclass{article}

\usepackage[final]{neurips_2025}

\usepackage[utf8]{inputenc} 
\usepackage[T1]{fontenc}    
\usepackage{hyperref}       
\usepackage{url}            
\usepackage{booktabs}       
\usepackage{amsfonts}       
\usepackage{nicefrac}       
\usepackage{microtype}      
\usepackage{xcolor}         
\usepackage{bm}
\usepackage{amsmath}
\usepackage{graphicx} 
\usepackage{xspace}
\usepackage{tcolorbox}
\usepackage[caption=false,font=footnotesize]{subfig}

\newcommand\nonumfootnote[1]{%
\begingroup%
    \renewcommand\thefootnote{}\footnote{\hspace{-3.7pt}#1}%
    \addtocounter{footnote}{-1}%
\endgroup%
}

\newcommand{\tocite}[1]{\textcolor{blue}{[TO CITE]}}

\newcommand{\ourmethod}{\textit{Omni-Video}\xspace}

\title{\ourmethod: Democratizing Unified Video Understanding and Generation} 

\author{
    Zhiyu Tan$^{1, 2\dagger}$ ~
    Hao Yang$^{2\dagger}$ ~
    Luozheng Qin$^{2}$  ~
    Jia Gong$^{2}$ ~
    Mengping Yang$^{2*}$ ~
    Hao Li$^{1,2*}$ \\[5pt]
    $^{1}$Fudan University \quad
    $^{2}$Shanghai Academy of Artificial Intelligence for Science \\[5pt]
    Project Page: \url{https://howellyoung-s.github.io/OmniVideo_project/}
}

\begin{document} 
\maketitle

\begin{abstract}
Notable breakthroughs in unified understanding and generation modeling have led to remarkable advancements in image understanding, reasoning, production and editing, yet current foundational models predominantly focus on processing images, creating a gap in the development of unified models for video understanding and generation.
This report presents \ourmethod, an efficient and effective unified framework for video understanding, generation, as well as instruction-based editing.
Our key insight is to teach existing multimodal large language models (MLLMs) to produce continuous visual clues that are used as the input of diffusion decoders, which produce high-quality videos conditioned on these visual clues.
To fully unlock the potential of our system for unified video modeling, we integrate several technical improvements:
1) a lightweight architectural design that respectively attaches a vision head on the top of MLLMs and a adapter before the input of diffusion decoders, the former produce visual tokens for the latter, which adapts these visual tokens to the conditional space of diffusion decoders;
and 2) an efficient multi-stage training scheme that facilitates a fast connection between MLLMs and  diffusion decoders with limited data and computational resources.
We empirically demonstrate that our model exhibits satisfactory generalization abilities across video generation, editing and understanding tasks.
Our code is publicly available at: \url{https://github.com/SAIS-FUXI/Omni-Video}.
\nonumfootnote{$\dagger$ Equal Contribution.}
\nonumfootnote{* Corresponding Authors.}
\end{abstract}
\begin{figure}[h]
    \centering
    \includegraphics[width=\linewidth]{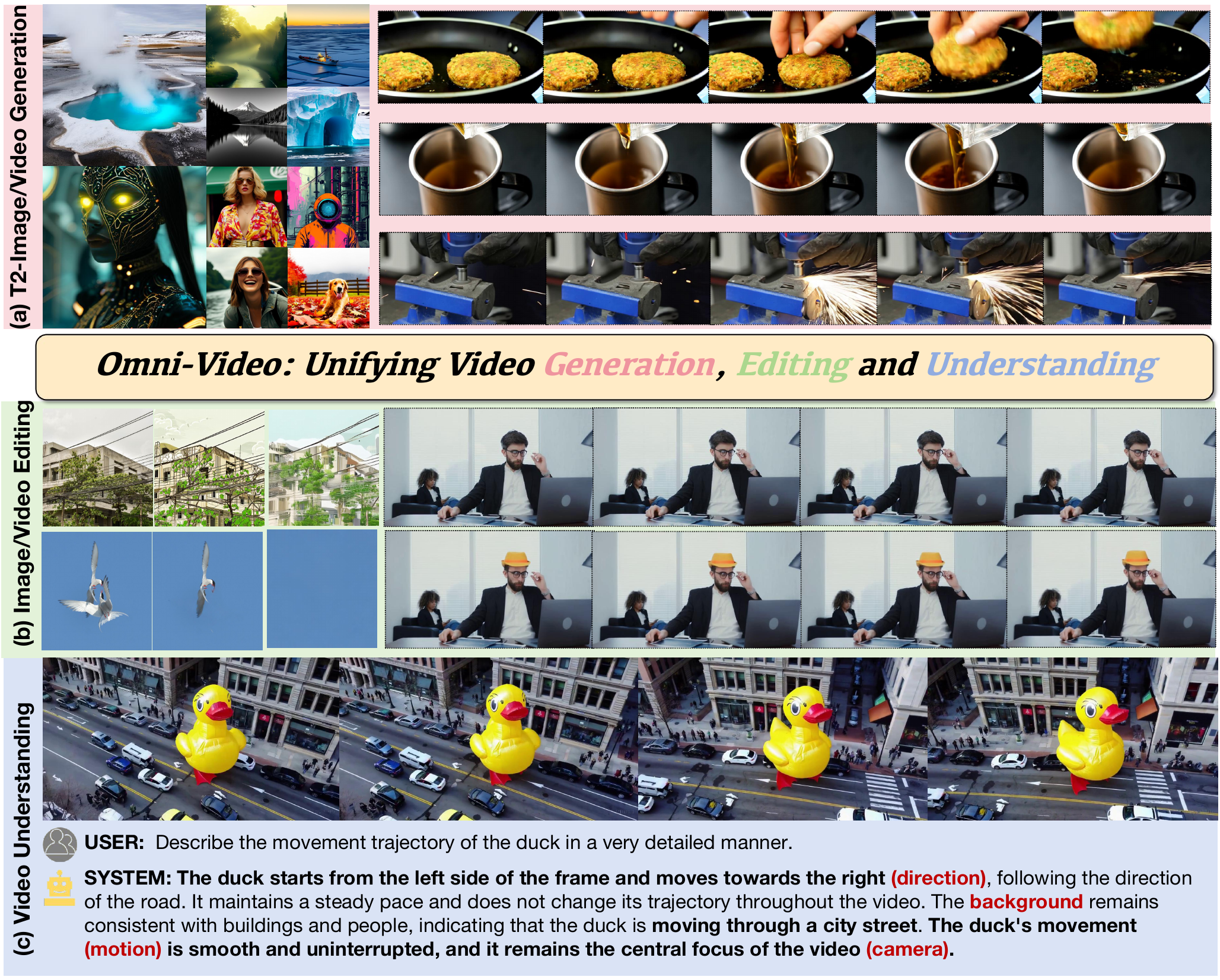}
    \vspace{-7pt}
    \caption{
    \textbf{Comprehensive illustration of the fundamental capabilities of our \ourmethod} on (a) photorealistic text-to-image/video generation, (b) image/video editing (\emph{i.e.,} change the style of source image toward target image and remove objects in the source image), and (c) video understanding (describe directions, background, motion, and camera movement, \emph{etc.}), all these tasks could be accomplished within a single unified architecture.
    }
    \label{fig:teaser}
\end{figure}
%

\section{Introduction}
The remarkable advancement of artificial intelligence in recent years, especially multimodal understanding~\cite{chen2024internvl, bai2025qwen25_vl} empowered by large-language models (LLMs) and multimodal generation\cite{dalle3, sdxl, wan2025wan} models driven by diffusion models, has enabled thrilling possibilities to go beyond existing experiences across various modalities, including vision~\cite{redmon2016you,rombach2022high,he2016deep,dosovitskiyimage}, language~\cite{devlin2019bert,brown2020language,raffel2020exploring}, as well as audio~\cite{baevski2020wav2vec,van2016wavenet,park2019specaugment,radford2023robust}, etc.
%
%
Despite these progresses, most understanding and generation systems are often designed separately, \emph{i.e.}, each specialized for a specific task or modality, leading to significant modeling redundancy and laborious computing complexity.
Alternatively, unified multimodal models~\cite{chen2025janus, tong2024metamorph} have emerged as a promising solution and gained increasing attention for their ability to input/output diverse data types seamlessly.
Such unified pipeline is trained jointly across multiple modalities to model task-agnostic representations~\cite{wang2025diffad,chou2018unifying} within a single framework, thereby removing tailored modality-specific components while achieving strong performance on both individual and compositional tasks~\cite{hu2021unit,wang2024git}.
%

Following this philosophy, recent efforts~\cite{chen2025blip3, deng2025bagel} have focused on developing unified architectures that simultaneously handle multimodal understanding and generation tasks.
Considering that the transformer-based autoregressive (AR) architecture serves as the predominant architectural paradigm for multimodal understanding , earlier approaches project both texts and images into discrete tokens to facilitate AR modeling across modalities~\cite{chen2025janus, wang2024emu3}.
For instance, Emu3~\cite{wang2024emu3} tokenize images, texts, and videos into a discrete space and train a single transformer model via next-token prediction on these multimodal sequences.
Chameleon~\cite{team2024chameleon} designed an early fusion scheme to project all modalities into a share discrete representational space for better interaction and reasoning.
Motivated by the great success of diffusion models~\cite{sdxl, dalle3} in visual generation, the most recent works~\cite{chen2025blip3, deng2025bagel} design hybrid diffusion-AR architectures that incorporate diffusion decoders to translate high-level visual tokens into photorealistic outputs.
Representative works such as Transfusion~\cite{zhou2024transfusion} and Show-O~\cite{xie2024show} respectively employ continuous and discrete diffusion modeling with full attention for improved visual quality.
%

However, training the above unified models from scratch requires substantial computational resources and a massive amounts of multimodal data.
Additionally, integrating MLLMs and diffusion models within a unified system poses significant challenges in architectural modifications and hyperparameter tuning.
To tackle these challenges, MetaMorph~\cite{tong2024metamorph} demonstrated that an improved visual understanding inherently endows MLLMs with visual generation capabilities. 
Building on this insight, a straightforward visual predictive instruction tuning (VPiT) strategy was introduced to train LLMs to generate continuous visual tokens, which are subsequently decoded into plausible images using a modified diffusion model.
Accordingly, MetaMorph achieved competitive performance on both image understanding and generation tasks with minimal architectural changes and computational overhead.
Nevertheless, previous unified models have predominantly targeted static images and show limited capability when it comes to processing videos, where accurately modeling temporal dynamics, ensuring motion coherence, and maintaining frame-wise consistency are essential.
Furthermore, the elevated data and computational demands for video modeling impose significant scalability constraints on current unified frameworks.
Consequently, delivering an appropriate framework for unified video understanding and generation is critically important for immersive media understanding, production, and even interaction.
%

To fill the vacant, this technical report proposes \ourmethod, an efficient and effective framework for unified video understanding and generation that combines a lightweight architectural design with a multi-stage training scheme.
Borrowing the exceptional video understanding and generation capabilities of existing MLLMs (\emph{e.g.,} VILA~\cite{lin2024vila}) and text-to-video (T2V) diffusion (\emph{e.g.,} Wan~\cite{wan2025wan}) models, our method aims at teaching existing video understanding MLLMs to produce video tokens that are `understandable' for existing T2V diffusion models.
More specifically, we first tuning MLLMs to produce textual tokens and visual tokens with separate text and vision heads following~\cite{tong2024metamorph}, the vision head generates continuous video tokens that is aligned with the representation of a pretrained vision encoder, \emph{e.g.,} SigLIP2~\cite{tschannen2025siglip2}.
Then, we adapt the continuous video tokens into the T2V diffusion model's conditional space with a dedicated lightweight adapter, such that the T2V model could synthesize faithful videos directly from the continuous video tokens.
Notably, only a small subset of model parameters, \emph{i.e.,} the text and vision head in stage-1 and the adapter in stage-2, are optimized.
%
To further activate the potential visual generation capability of MLLMs, we unlock the parameters of text and vision head, adapter, as well as the T2V model and train the whole system with meticulously curated high-quality understanding and generation data, thereby improving the synthesis quality and the understanding ability.
%
This selective fine-tuning strategy requires only minimal data and computational resources, allowing for efficient adaptation without compromising the foundational performance of the pretrained model.

In addition to generate harmonious videos conditioned on user instructions, we also extend our unified framework towards detail-preserving visual editing scenarios.
One straightforward solution is to feed original signals into the visual encoder of MLLMs and retrain the unified model to reconstruct unedited areas and penalize editing areas with paired editing data\cite{yu2024promptfix, zhao2024ultraedit, zhang2024hive}.
This approach, however, incurs substantial computational overhead with massive data and often leads to suboptimal performance as MLLMs primary focus on high-level semantic features rather low-level details.
Since T2V diffusion models are trained to recover clean videos from corrupted signals, they inherently excel at processing fine-grained visual information.
Accordingly, we leverage pretrained VAEs to extract low-level compressed priors from original input and directly send them to the diffusion model.
Such that, the original signals interact with noised embeddings throughout all denoising steps, ensuring high-fidelity visual preservation during the editing process.
%
%
Furthermore, motivated by the recent chain-of-thought reasoning capabilities in the LLMs community, we devise a ``think mode" to achieve higher-quality visual outputs through logical inference and contextual planning in video generation and editing.
Together, our framework enables seamlessly predicting multimodal tokens for video understanding, generation, as well as video editing tasks, with minimal architectural modification, diverse data sources, and computation-efficient optimization.

In order to testify the effectiveness of our proposed \ourmethod, we conduct extensive experiments across various tasks including image/video understanding, text-to-image/video generation, image/video editing (Fig.~\ref{fig:teaser}).
The results demonstrate that our method achieves favorable results on these tasks within a unified model.
%
Notably, the integration of a chain-of-thought reasoning module (``think mode") yields better understanding of users' instructions and obtains consistent 
improvements in output quality, suggesting the great potential of inference time scaling in unified models.
To sum up, our primary contributions are summarized blow:
\begin{itemize}
    \item \textbf{A unified paradigm for video unified modeling:} We introduce \ourmethod, a novel framework that integrates video understanding and generation. By teaching the multimodal understanding model to produce continuous vision tokens for diffusion to generate realistic videos, our model successfully tackles complex video understanding and generation within a single model.
    \item \textbf{An efficient multi-stage training recept for resource-friendly optimization:} We devise a mutl-stage training scheme that accomplish 1) fast alignment between MLLMs' visual tokens and pre-trained visual encoder, and 2) efficient adaptation of the visual token toward T2V models' conditional space. Accordingly, our model is effectively trained without massive computation resources.
    \item \textbf{A thorough empirical evaluation demonstrates superior performance:} Extensive experiments under various video understanding and generation scenarios demonstrate that \ourmethod achieves prevailing performance. Moreover, the integration of video/image editing enhances the model's flexibility as a general-purpose framework.
\end{itemize}

\section{Related Work}

\subsection{Multimodal Understanding}
Taking advantage of advanced large language models (LLMs)~\cite{achiam2023gpt4, grattafiori2024llama3, yang2025qwen3}, multimodal large language models (MLLMs) have gained extensive research attention and demonstrated magnificent capabilities in understanding and reasoning multimodal content~\cite{li2024llava-ov, bai2025qwen25_vl, guo2025seed15_vl}.
In order to inject knowledge beyond the text modality into LLMs, existing methods typically employ modality-specific encoders (\emph{e.g.,} CLIP~\cite{radford2021clip}) to extract latent embeddings and perform alignment within the LMMs' transformer space.
For instance, LLAVA~\cite{liu2023llava, liu2024llava_1_5} and MiniGPT-4~\cite{zhu2023minigpt4} connect pre-trained visual encoders and LLMs with simple projection layers, enabling LLMs to perform complex multimodal understanding while preserving their textual reasoning ability.
Building this insights, more advanced techniques such as scaling the visual encoders~\cite{chen2024internvl}, support dynamic and high input resolutions~\cite{wang2024qwen2-vl}, employment of Mixture-of-Experts (MOE) architectures~\cite{jiang2024mixtral, liu2024deepseekv2}, replacing linear projection layers with different architectures~\cite{li2023blip2, alayrac2022flamingo}, improvement of training receipts~\cite{karamcheti2024prismatic, lin2024vila} are developed to enhance the fine-grained perception capabilities of MLLMs.
Similarly, video-centric MLLMs~\cite{lin2024vila} encode videos into sequential visual tokens that are compatible with LLMs.
Although these approaches excel at multimodal understanding and reasoning, they primarily focus on perception tasks and lack the capability of producing other modalities beyond text.

\subsection{Multimodal Generation}
%
Early multimodal generation models employ generative adversarial networks (GANs)~\cite{goodfellow2014gan} to synthesis novel outputs. However, the scalability, visual quality and resolution of GANs are relatively poor.
Diffusion models (DMs)~\cite{DDPM} and autoregressive (AR) models~\cite{box1970time, radford2018gpt}, equipped with their inherent stable optimization and scalability, the transformer architecture, large-scale training and infrastructure, and trillion-level multimodal corpus, are current state-of-the-art for image/video synthesis.

\noindent \textbf{Diffusion models (DMs).}
DMs gradually transform simple noise distributions into complex data distributions through a series of learned denoising steps.
The pioneering work laid the theoretical foundations, while numerous subsequent studies have enhanced the efficiency and output quality through improved training/sampling strategies~\cite{lu2022dpm, karras2024analyzing, sdxl, gao2025seedream} and optimization techniques~\cite{lu2024simplifying, song2023improved, liu2024playground}, \emph{etc.}
In particular, the introduction of stable diffusion (SD) and diffusion transformer (DiT) ~\cite{peebles2023dit} directly caused the blooming of high-quality, high-resolution, and long-duration visual generation.
For instance, Sora~\cite{openai2024sora}, built on the diffusion transformer architecture, achieved superior performance in producing detailed, high-resolution videos that adhere to the user’s instructions.
Follow-up approaches including open-sourced\cite{zheng2024opensora, tan2025raccoon, wan2025wan} and closed-source models~\cite{klingai2024, gao2025seedance} have further advanced the state-of-the-art in text-to-video generation.
These developments underscore that enhanced diffusion techniques are enabling richer, more dynamic, and semantically coherent visual content creation.
In this work, we integrate the open-sourced Wan2.1~\cite{wan2025wan} video generation model into our unified understanding and generation system as its leading performance in the open-source community.

\noindent \textbf{Autoregressive (AR) models.}
Borrowing the success of AR modeling in language domain, another line of approaches learn to synthesize visual content in autoregressive manner.
Concretely, visual inputs are first mapped into a sequence of discrete tokens, after which both visual and text tokens are modeled via next token prediction.
 To encode continuous visual signals into discrete tokens, vector quantization techniques\cite{yu2023magvitv2, tian2024var, li2024mar} are usually applied to high-dimensional data into a finite set of latent codes (codebook).
DALLE~\cite{dalle2, dalle3} demonstrated that scaling with sufficient data autoregressively could generate high-quality images from textual descriptions.
MAGVIT-v2~\cite{yu2023magvitv2} showed that GPT-style LLMs might outperform diffusion models on visual generation benchmarks with an improved video tokenizer.
With this top-performing tokenizer, VideoPoet further identified the potential of AR model, which was trained on discrete visual, text and audio latents, in producing plausible videos, especially excel in modeling complex large motions.
Differently, VAR~\cite{tian2024var} designed a coarse-to-fine ``next-scale prediction" paradigm to learn visual distributions and MAR~\cite{li2024mar} applied AR modeling in the continuous space by modeling per-token distribution by diffusion.

\subsection{Unified Multimodal Understanding and Generation}
In an effort to build a single system for both understanding and generation tasks, substantial developments have been witnessed in unified multimodal models recently.
In addition to the text modality, these models process various modalities as input (\emph{e.g.,} image, video) and also output various modalities in a unified manner.
Depending on whether visual signals are encoded into discrete or continuous tokens, existing unified multimodal models primarily consist of pure AR models~\cite{chen2025janus, wang2024emu3, wu2025liquid} and diffusion-AR hybrids~\cite{tong2024metamorph, chen2025blip3, liao2025mogao}.
For example, both Chameleon~\cite{team2024chameleon} and Emu3~\cite{wang2024emu3} represented image/videos with VQ-VAE~\cite{van2017vqvae} as discrete tokens and trained uniform transformer-based LLM models from scratch in next token prediction manner.
Janus and Janus-pro~\cite{chen2025janus} decoupled visual encoding with SigLIP~\cite{zhai2023siglip} and VQ-VAE~\cite{sun2024llamagen} for multimodal understanding and generation, respectively.
Such decoupling ameliorated the conflict of visual encoder for understanding and generation tasks.
Following this philosophy, the most recent work converted images into discrete tokens with a text-aligned tokenizer that was aligned with LLM embeddings.
However, discreting visual signals often struggles to preserve fine-grained details, leading to sub-optimal synthesis performance.
Consequently, UniFluid~\cite{fan2025unifluid} build a unified framework from continuous visual tokens following the MAR's per-token distribution modeling.

Alternatively, other approaches integrate diffusion models into LLMs for visual decoding.
In particular, Show-O~\cite{xie2024show} and Tranfusion~\cite{zhou2024transfusion} unified diffusion models for generation tasks and LLMs for textual responses, while the former adopted discrete diffusion and the latter used continuous diffusion.
Along this line, Show-O2~\cite{xie2025show2} and Mogao~\cite{liao2025mogao} improved the model performance via combining AR modeling and flow-matching~\cite{lipman2022flow}.
In order to identify the impact of core components (\emph{i.e.}, image representation, training objective, training strategies) of hybrid diffusion-AR unified models, BLIP3-o~\cite{chen2025blip3} presented a thorough empirical study.
Despite these models' success in unifying understanding and generation, training these models from scratch requires massive compute and data resources.
The most related work to ours is MetaMorph, which trained LLMs to predict multimodal tokens for generation with visual-predictive instruction tuning (VPiT).
However, most of existing methods, including MetaMorph, focus on modeling static images and show very limited ability for processing videos.
Therefore, this work presents an efficient and effective framework for unified video understanding and generation.
%

\section{Model Architecture}
\label{sec:model_architecture}

\begin{figure*}[t]
  \centering
  \includegraphics[width=1\textwidth]{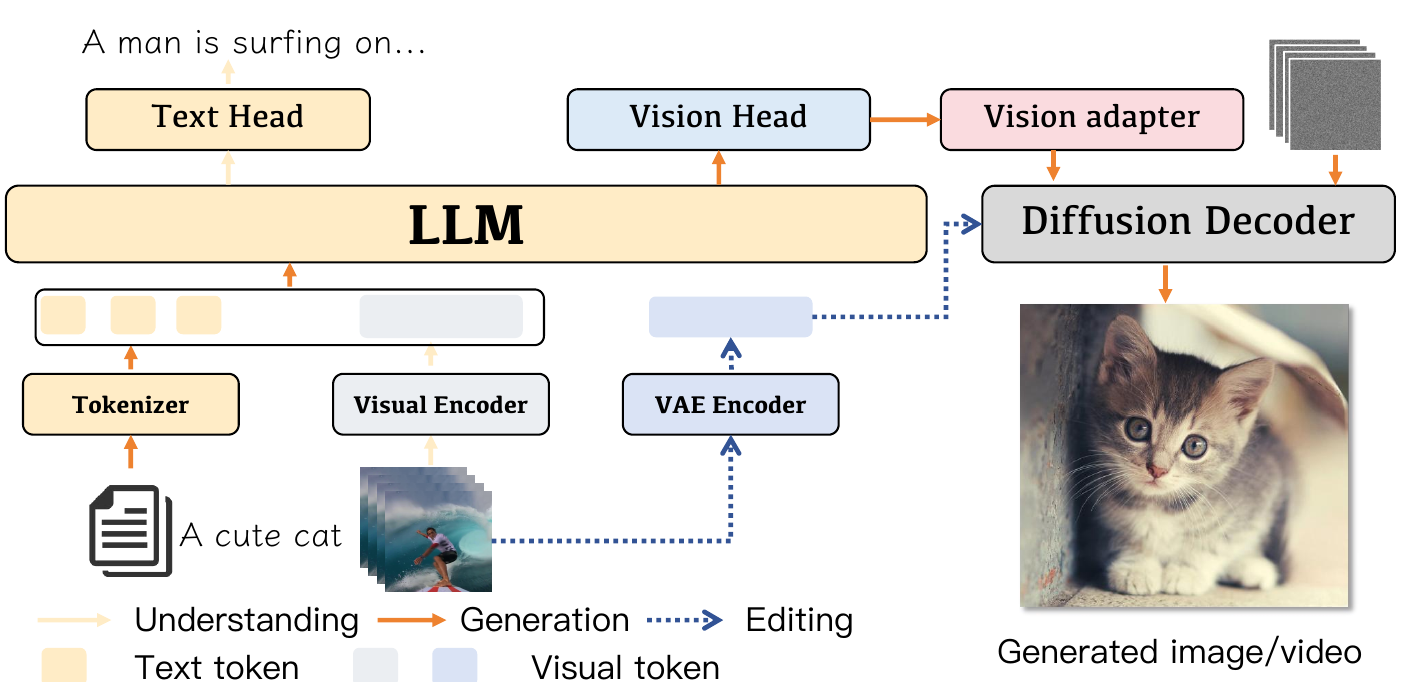}
  \caption{
  \textbf{Overall architecture of our proposed \ourmethod for unified video understanding and generation.}
  We connect MLLMs' exceptional understanding capability with the visual generation ability of diffusion decoders with lightweight architectural design, enabling MLLMs to produce visual continuous tokens that are decoded into photorealistic images/videos by the diffusion decoder.
  }
  \label{fig:model_arch}
\end{figure*}

\subsection{Overview}
Recall that our aim is to build a unified architecture that enables multimodal video understanding and generation simultaneously. 
To achieve this, we teach multimodal understanding models (\emph{e.g.},VILA~\cite{lin2024vila}) to produce visual tokens as the input of a text-to-video diffusion model with lightweight adapters and a novel three-stage training scheme.
In this way, the alignment between the output visual tokens and the diffusion model's context conditions could be achieved efficiently, without requiring training the whole system from scratch.
Fig.~\ref{fig:model_arch} presents the overall structure of our \ourmethod. 
Overall, our model connects an AR-based MLLM model for understanding and a DiT-based diffusion model for visual generation, more details are presented below.

\subsection{Producing Visual Tokens with MLLMs}
Prior MLLMs accept both textual tokens and visual embeddings but only produce textual tokens, to support both textual and visual outputs within a unified MLLM model, we design two distinct prediction heads, \emph{i.e.,} one \textbf{Text Head} for textual output and one \textbf{Vision Head} for visual output.
Formally, given a multimodal input consisting of $M$ textual tokens \(T = \{t_1, t_2, \dots, t_M\}\), with each token \(t_i \in {T}\) drawn from a fixed vocabulary \(\mathcal{T}\), and $N$ visual content represented as embeddings \(V = \{\mathbf{v}_1, \mathbf{v}_2, \dots, \mathbf{v}_N\}\), where each embedding \(\mathbf{v}_t \in \mathbb{R}^{D}\) represents embeddings of one frame extracted from pre-trained visual encoders.
Our unified MLLM model seeks to jointly model this combined input sequence \(I = [T; V]\) and producing \(O = [T_o; V_o]\), where \(T_o\)\ and \(V_o\)\ respectively represent textual tokens from the text head and visual outputs from the vision head.

More specifically,  the text token performs standard language modeling, predicting tokens from a vocabulary augmented with four newly introduced special tokens: \texttt{\textless BOI\textgreater}, \texttt{\textless EOI\textgreater}, \texttt{\textless BOV\textgreater}, and \texttt{\textless EOV\textgreater}. 
Conversely, the vision head maps the MLLMs' hidden states into a continuous visual tokens.
To enforce strong semantic grounding and stable latent predictions, the vision head is trained to align the continuous visual embeddings extracted by off-the-shelf visual encoders (\emph{e.g.}, SigLIP-v2\cite{tschannen2025siglip2}).
Such alignment can be accomplished by computing the similarities (\emph{e.g.,} L2 distance, cosine similarity) between the visual tokens $V_o$ from vision head and visual embeddings $E$ from the pre-trained visial encoders: 
\begin{equation}
    \mathcal{L}_{vision} = \| V_o - E\|^2.
\end{equation}
Regarding the text head, it aims to predict text sequences \( \hat{T} = [\hat{t}_1, \hat{t}_2, \ldots, \hat{t}_{m'}] \) from the input of \( T = [t_1, t_2, \ldots, t_m] \).
This sequence is optimized according to the cross-entropy loss defined as:
\begin{equation}
    \mathcal{L}_{text} = -\sum_{j=1}^{m} \log P(\hat{t}_j | t_j).
\end{equation}
Once trained, the two distinct heads are separately responsible for generating textual outputs for understanding tasks and visual tokens for generation tasks.
Such explicit modality-marking modality-marking strategy not only resolves ambiguity but also significantly enhances the coherence and fidelity of multimodal generation.
Notably, only the vision head and the text head are trained, which maintains the original multimodal understanding capabilities of MLLMs to a great extent and efficiently turns understanding-only MLLMs to generate visual clues.

\subsection{Making Visual Tokens Understandable for DMs}
After finetuning a pre-trained MLLM to synthesize both visual and text tokens with separate heads, we turn to decoding the output visual tokens as photorealistic images and videos.
Inspired by the great success of diffusion models (DMs) in the context of image and video generation, we incorporate pre-trained T2V diffusion models as our visual decoder.
Along this line, we design a dedicated lightweight adapter to project the visual context sequences $V_o$ from the vision head into a compact embedding space that is ``understandable" for the diffusion model.
Formally, given the output $V_o$ tokens, the adapter process them as:
\begin{equation}
    Q = \text{Adapter}(V_o),
\end{equation}
where $Q$ denotes the output of the adapter and is further adopted as the input condition of the diffusion decoder, guiding the diffusion decoder to generate images that are harmonious to the input instructions.

To make sure that the diffusion decoder could produce high-quality visual contents conditioned on $Q$, we devise an effective training strategy:
we first explicitly align the adapter's outputs with the conditional space (\emph{i.e.,} the textual embedding derived from text encoders) of the diffusion decoder, thus bridging the domain gap between multimodal context representations and the diffusion decoder's conditioning space.
Then, we train the adapter with real-world paired text and viusal images/videos with the diffusion decoder's training objective:
\begin{equation}
    \mathcal{L}_{\text{DMs}}(\theta) = E_{(X_1,Q)\sim{D}, t\sim{U}(0,1),X_0\sim{N}(0,1)} \left[ \|V_\theta(X_t,Q,t) - V_t\|^2 \right],    
\end{equation}
where $\theta$ denote the model's parameters, it is noteworthy that only the lightweight adapter is updated here, thus $\theta$ is the adapter's parameters.
${V}_\theta({X}_t,Q,t)$ denotes the predicted velocity based on an instance $(X_1,Q)$, timestep $t$, and noise $X_0$, and $Q$ is the output of the adapter.
This alignment significantly enhances the semantic consistency of the generated outputs while enabling efficient and stable training.
\subsection{Unifying Understanding, Generation and Editing}
In addition to supporting visual generation tasks, we further integrate the challenging video editing tasks into our unified framework.
Video editing requires the model to reliably understanding user instructions and precisely editing specific areas while preserving other details.
To achieve this, we introduce two generation modes for video generation and editing based on their conditional inputs.
In the generation mode, the adapter exclusively processes latents predicted by the vision Head along with corresponding textual embeddings, then the diffision decode produce videos conditioned on th adapter's output.
In contrast, when operating in editing mode with source condition videos available, we enhance the input to the adapter by incorporating sequence embeddings of these source videos. These embeddings are extracted using a visual compressor, specifically a 3D-causal-VAE.
In this way, we can effectively integrate the spatial and temporal features of the source videos into the editing process. 
%

%
%

Concretely, for each sample, we concatenate the following modalities into a unified conditional embedding sequence:
$\mathbf{c} = [\mathbf{v}_1, \dots, \mathbf{v}_T \parallel \mathbf{h}_1, \dots, \mathbf{h}_{T'} \parallel \mathbf{t}_1, \dots, \mathbf{t}_m]$,
where $\mathbf{v}_t$ represents the vision latent predictions from the AR Vision Head, and $\mathbf{h}_t$ are optional spatial-temporal embeddings derived from a pretrained 3D-Causal-VAE utilized specifically in editing scenarios. 
Moreover, $\mathbf{t}_1$ denotes the original textual conditional input of the diffusion decoder.
The conditional sequence $\mathbf{c}$ is then projected via the adapter module and subsequently injected into every block of the DiT via cross-attention mechanisms.
Such explicit conditioning ensures that the diffusion model accurately adheres to the intended semantic content and retains spatial-temporal consistency across frames.
During the training process, the system automatically switches between these two modes in response to generative and editing data inputs. 
Such adaptive mechanism ensures that the model can efficiently adjust its operational mode to optimize both generation and editing performance, thereby enhancing the overall effectiveness of the model in handling diverse data types.
\section{Data and Training}

\begin{table}[t]
    \centering
    \caption{Summary of datasets used in the joint vision and language pretraining stage.}
    \begin{tabular}{lll}
        \toprule
        \textbf{Category}       & \textbf{Dataset} & \textbf{Ratio} \\
        \midrule
        Image Understanding     & Total            & 37.9\% \\
        & JourneyDB-4M~\cite{sun2023journeydb}   & \\
        & LLaVA-v1.5~\cite{liu2024llava_1_5}            & \\
        & ShareGPT4V~\cite{chen2024sharegpt4v}            & \\
        & Cambrain-7M~\cite{tong2024cambrian}           & \\
        & MMC4~\cite{zhu2023mmc4}        & \\
        \midrule
        Video Understanding     & Total            & 4.36\% \\
        & LaVA-Video-178K~\cite{zhang2024llava-vid}       & \\
        & VSTaR-1M~\cite{zohar2024vstar}              & \\
        & ShareGPT4Video-40K~\cite{chen2024sharegpt4video}    & \\
        \midrule
        Text-to-image Generation& Total            & 33.05\% \\
        & LAION-5B~\cite{schuhmann2022laion}              & \\
        & COYO-700M~\cite{kakaobrain2022coyo-700m}             & \\
        \midrule
        Text-to-video Generation& Total            & 3.97\% \\
        & HD-VILA~\cite{xue2022hdvila}               & \\
        \midrule
        Text-to-image Editing   & Total            & 18.95\% \\
        & AnyEdit~\cite{yu2024anyedit}               & \\
        & UltraEdit~\cite{zhao2024ultraedit}             & \\
        & HIVE~\cite{zhang2024hive}                  & \\
        & PromptFix~\cite{yu2024promptfix}             & \\
        \midrule
        Text-to-video Editing   & Total            & 1.76\% \\
        & Senorita-2M~\cite{zi2025senorita}           & \\
        \bottomrule
    \end{tabular}
    \label{tab:datasets}
\end{table}

\begin{figure*}[htp]
  \centering
  \includegraphics[width=0.8\textwidth]{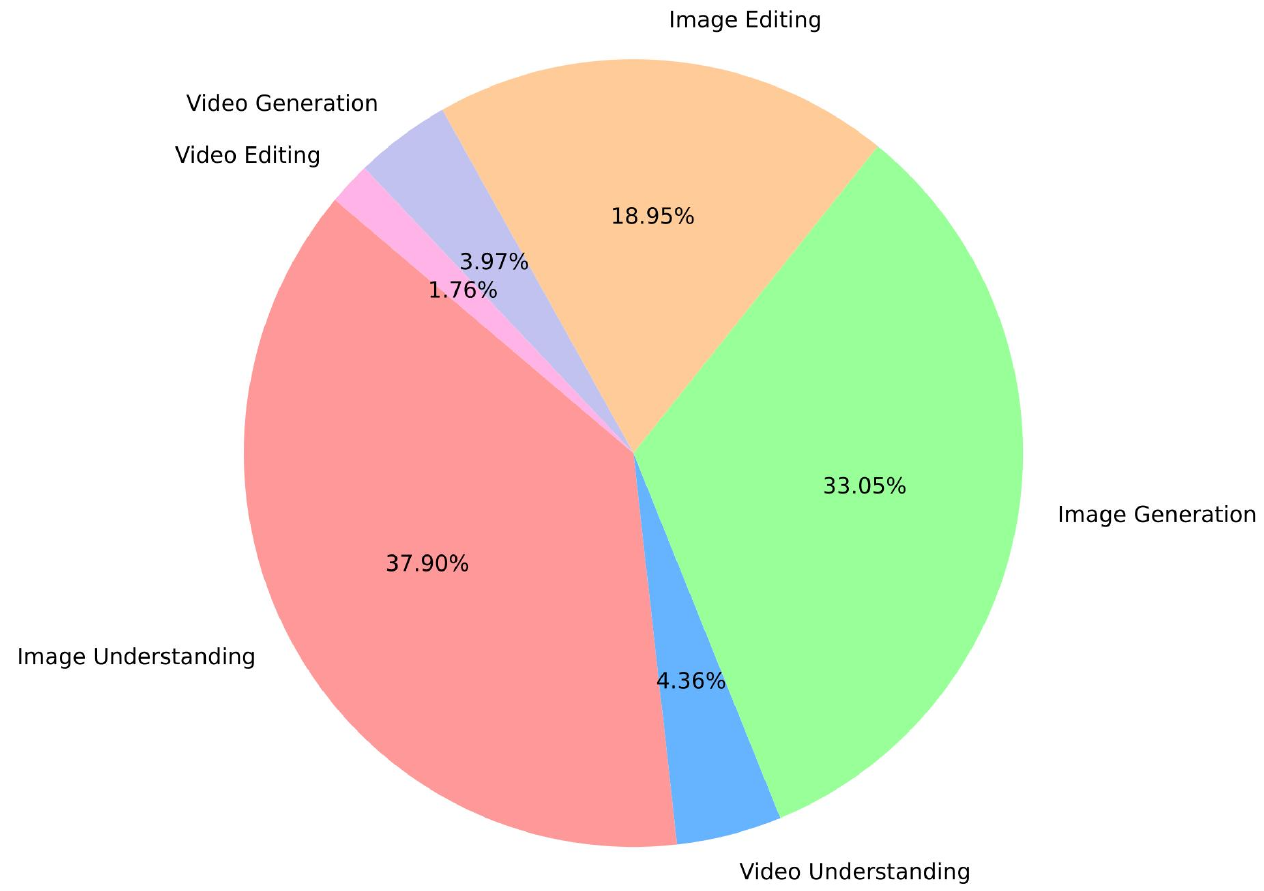}
  \caption{
  \textbf{Distribution of datasets used in the joint vision and language pretraining stage.}
  }
  \label{fig:datasets}
\end{figure*}

\subsection{Data}
To train our \ourmethod, we use a mixture of open-source multimodal data including image understanding, video understanding, text-to-image, text-to-video, instruction-based image and video editing data.
Tab.~\ref{tab:datasets} and Fig.~\ref{fig:datasets} present the detailed data sources used in our training.
In general, our dataset comprises approximately {17.2 million} samples for image understanding, {1.98 million} for video understanding, {15 million} for image generation, {8.6 million} for image editing, {1.8 million} for video generation, and {0.8 million} for video editing. 
This sufficient distribution highlights a strong emphasis on image-centric tasks while still providing sufficient resources to support key developments in video-related research.
For video understanding data, we use the original data format released by the official.
For the text-to image and text-to-video data, we conduct a comprehensive data curation pipeline to obtain high-quality training data.
The curation pipeline including but is not limited to:
quality filtering via assessing the resolution, ratio of width and height, aesthetic prediction, video clipping;
content filtering via evaluating watermarks, optical flow-based motion prediction, text overlays, OCR detection, \emph{etc.}
Moreover, the text conditions are re-captioned based on the visual content using a multimodal larga language model.

\subsection{Three-Stage Training Strategy}

\subsection{Training}
\begin{figure}[t]
    \centering
    \includegraphics[width=\linewidth]{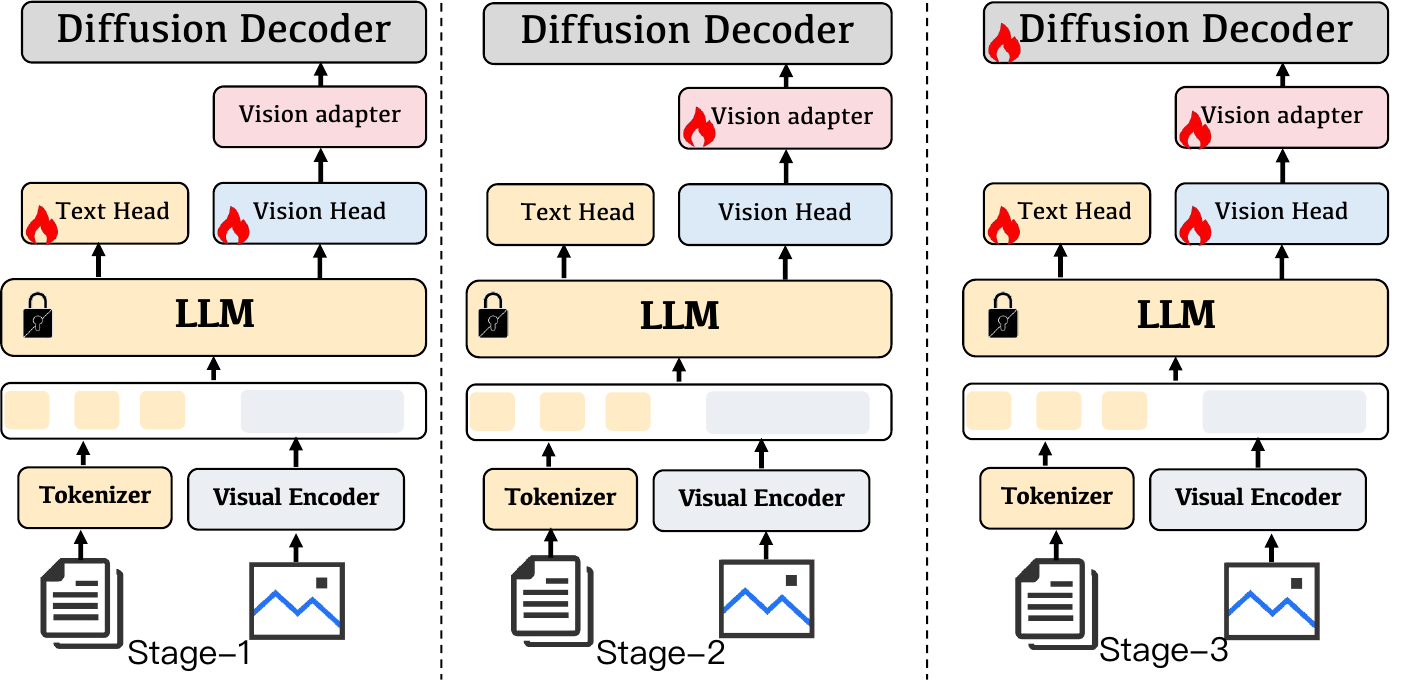}
    \caption{
    \textbf{Illustration of our multi-stages training strategy.}
    The flames denote the module parameters are trainable at each stage, and the block indicates that the model parameters are frozen.
    }
    \label{fig:multi_stages_strategy}
\end{figure}

As shown in Fig.~\ref{fig:multi_stages_strategy}, we employ a multi-stage training strategy to effectively connect multimodal understanding models (MLLMs) with text-to-video diffusion models, transforming MLLMs to to unified multimodal model.
To be more specific, we first attach a separate vision head on the top of MLLMs to produce visual tokens, followed by finetuning a lightweight adapter to align the visual tokens with the conditional space of the diffusion decoder.
Finally, we perform joint finetuning on both understanding blocks and generation blocks for high-quality visual generation.

\paragraph{Model Initialization.}
To mitigate the substantial training costs, we initialize the components of our unified model using readily available pretrained weights. 
Training either a Multimodal Large Language Model (MLLM) for video understanding or a diffusion model for video generation from scratch necessitates vast computational resources and data. Therefore, the selection of appropriate pretrained models for initialization is a critical step.
For the MLLM component, we adopt the weights from VILA~\cite{lin2024vila}. 
Regarding the diffusion model component, we initialize our model with the weights of the Wan 2.1 text-to-video model~\cite{wan2025wan}. This model was chosen for its compact parameter count of only 1.3B, which still provides a strong foundation for text-to-video generation. %
We clarify that the original Wan 2.1-1.3B model does not natively support visual editing capabilities, and we enable the diffusion decoder to perform image and video editing tasks via the proposed method mentioned earlier.
Moreover, the newly introduced adapter and vision head are initialized using a standard random initialization manner.

\paragraph{Stage-1: Teaching MLLMs to generate visual continuous tokens.}
In the initial stage, we keep the parameters of the MLLM fixed, focusing solely on training the text head and vision head, with the parameters of the diffusion model also remaining unchanged.
The trained loss includes:
1) the similarity loss between the visual tokens from the vision head and visual embeddings from the pre-trained visual encoders (Eq.1);
and 2) the cross-entropy loss that perform next token prediction on the text sequences (Eq.2).
The primary goals at this stage are twofold: 
Preserving the understanding capabilities of MLLMs and swiftly training the newly introduced components (\emph{i.e.,} the vision head) to produce visual continuous tokens.
During this phase, we incorporate data relevant to both understanding and generation tasks. 
As a result, the MLLM not only retains its video comprehension abilities but also learns to distinguish between understanding and generation tasks.
When the user instructions are to generate images/videos, the vision head produce tokens between the special tokens: \texttt{\textless BOI\textgreater}, \texttt{\textless EOI\textgreater}, \texttt{\textless BOV\textgreater}, and \texttt{\textless EOV\textgreater}, which are further used as input for the diffusion decoder.
%


\paragraph{Stage-2: Projecting visual continuous tokens into diffusion models' conditional space.}
In the second stage, we keep all understanding-related parameters fixed and focus exclusively on training the generative modules.
During this phase, we train the adapter to improve the alignment of semantic information from the understanding module with the latent space of the diffusion model.
Concretely, the output visual continuous tokens from the vision head are fed to the adapter, the adapter is first trained to align its output with the textual embedding space of the diffusion decoder.
Then, conditioned on the output of the adapter, we incorporate diffusion decoder into the training process and only update the adapter's parameters with the flow matching objective (Eq.(4)).
To fully unlock the potential of the diffusion decoder for various geneartion tasks, we employ a diverse dataset comprising four different tasks: text-to-image, text-to-video, image editing, and video editing.
Originally, the model was constrained to text-to-video synthesis.
However, our multi-task training approach equips it with a broader range of capabilities, demonstrating enhanced generalization abilities.
%


\paragraph{Stage-3: Jointing finetuning for high-quality visual generation.}
To further improve the quality of generated visuals, we unfreeze and simultaneously fine-tune the vision head, adapter, and the diffusion decoder. 
During this stage, we incorporate higher-quality training data and increase the video sampling rate from 8 frames per second (fps) to 12 fps. 
Such adjustment is particularly advantageous for generating videos with substantial motion. 
Despite the increase in trainable parameters and a 50\% rise in target frames, this phase does not significantly escalate the overall training cost. 
This efficiency is attributed to the solid generative foundation established in the previous stages, facilitating the effective optimization of generation quality.

\paragraph{Multi-task mixed training strategy.}
As mentioned earlier, our strategy involves multi-task training. While this approach can enhance generalization, it often slows down the convergence. 
To address this, we employ two acceleration strategies:
\begin{enumerate}
    \item \textbf{Two-sub-stage multi-task Learning:} We organize the multi-task training into two distinct sub-stages. 
    In the first sub-stage, training is focused exclusively on the text-to-image and text-to-video tasks. This method achieves two key objectives: it rapidly develops the model's image generation capabilities and establishes a strong foundation for the later image editing and video editing tasks.
    In the second sub-stage, we incorporate all four tasks into joint training. By commencing the editing tasks from a well-initialized foundation, we facilitate faster convergence.
    
    \item \textbf{Hybrid Mixed-Batching:} 
    We employ a hybrid batching strategy tailored for different tasks and datasets. 
    The training data is divided into two main groups: (1) video generation and editing, and (2) image generation and editing. 
    For each group, batches are constructed separately, ensuring that data with the same resolution and frame count are processed together in a single forward and backward pass. This approach enhances computational efficiency. In contrast, a common alternative is to create separate batches for each individual task. 
    Although this also permits multi-task learning within a single iteration, it can decrease the frequency of parameter updates for a given level of throughput (thereby maintaining high GPU utilization). Our hybrid strategy sustains high throughput while increasing the frequency of model updates, leading to faster convergence.
\end{enumerate}

\paragraph{Think Mode.}
To endow our model with generative reasoning capabilities, we have developed a straightforward "thinking" mode. In this mode, the MLLM first performs a "reasoning rewrite" on the input prompt. Concurrently, it adjusts the output of the Vision Head to align with the rewritten prompt. Consequently, the diffusion model can generate an image or a video based on this new rewritten prompt and its corresponding Vision Head output. This thinking mode effectively leverages the advanced reasoning capabilities of the MLLM model to achieve reasoning-based generation, all without necessitating any retraining of the diffusion model.
%

\paragraph{Memory optimization and acceleration.}
We employ several techniques to optimize memory usage and expedite training.
To minimize the memory footprint, we utilize DeepSpeed's optimizer partitioning along with the ZeRO Stage-1 strategy. 

Additionally, our multi-task configuration leads to longer conditioning sequences due to the inclusion of tokens from a 3D VAE encoding of the source video. 
To manage this, we use sequence parallelism to distribute the memory load across multiple GPUs. 
Lastly, gradient checkpointing is implemented to reduce memory consumption from activations, especially when training with higher-resolution data.
\section{Experiments}

\begin{figure*}[t]
  \centering
  \includegraphics[width=1\textwidth]{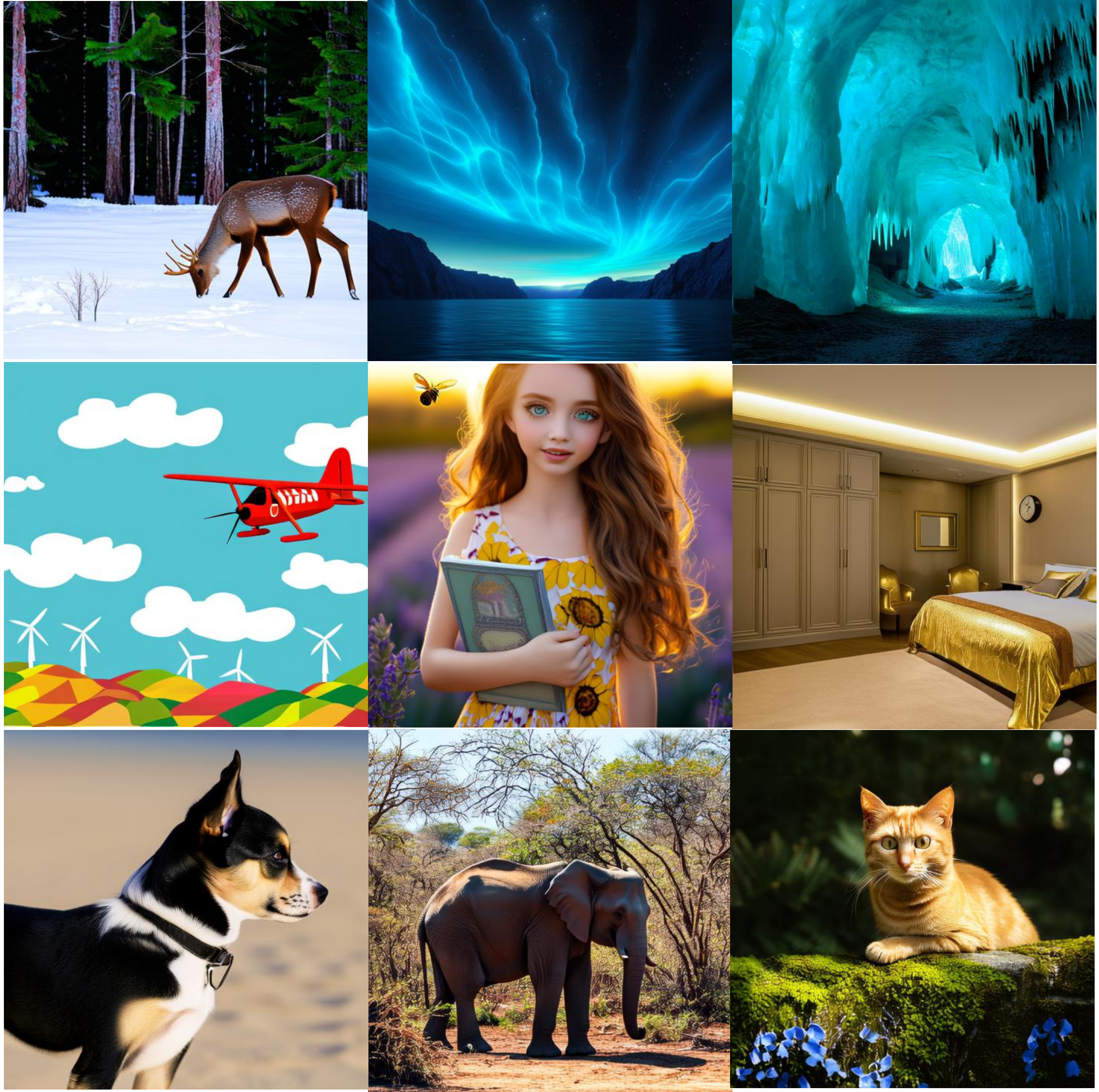}
  \caption{
  \textbf{Example images generated by our proposed \ourmethod on text-to-image settings.}
  }
  \label{fig:t2i}
\end{figure*}

\subsection{Experiment Settings}
We employed distinct hyperparameter configurations for different tasks. For all generative tasks, we adopted the UniPC sampling strategy and utilized classifier-free guidance (CFG).
The CFG scale is set as $5.0$ for text-to-image generation tasks and $3.0$ for text-to-video generation tasks, and the sampling steps are $50$ for the former and $40$ for the latter.
Additionally, we use a universal negative prompt instead of a null condition to enhance output quality across all generative tasks.
Regarding visual understanding, a top-p sampling strategy of 0.6 was applied to control diversity, alongside a temperature of 0.2 to manage randomness. The maximum number of new tokens generated was set to 1024.
Moreover, the prompts used in our evaluation are given in the appendix.

\subsection{Experiment Results}
\paragraph{T2I performance on GenEval benchmark}
We evaluated our model's visual generation performance on the GenEval benchmark. It is crucial to note a significant discrepancy between our training data and the prompts used in GenEval. Our model was trained on prompts rich in detail, whereas GenEval exclusively uses very short prompts for testing. We did not perform any special fine-tuning to address this domain gap. Despite this, our model achieved performance comparable to most state-of-the-art unified models on GenEval, as shown in ~\ref{tab:geneval_comparison}. It is important to emphasize that previous unified models have primarily focused on the understanding and generation of images, whereas our work extends this paradigm to the video domain.

\paragraph{T2V performance on VBench-Long benchmark.}
We evaluated our model’s long-range video generation performance on the VBench-Long~\cite{huang2024vbench} benchmark, which comprises 16 fine-grained metrics. As shown in Tab.~\ref{tab:vbench_forced}, our method achieves state-of-the-art results in \textit{Subject Consistency} (98.39\%), \textit{Temporal Flickering} (99.87\%), \textit{Object Class} recognition (93.54\%), and \textit{Temporal Style} consistency (25.81\%). These results highlight the model’s strong ability to preserve temporal coherence, maintain subject identity, and capture semantic details over extended sequences. Compared with Wan2.1-T2V-1.3B, which reports the highest quality score (85.23\%), our model delivers competitive overall performance while surpassing it in 10 out of 16 categories, demonstrating the robustness and generalization ability of our unified video generation framework.

\begin{table}[t]
\centering
\caption{\textbf{T2I performance comparison on GenEval.}}
\label{tab:geneval_comparison}
\resizebox{\linewidth}{!}{%
\begin{tabular}{lccccccc}
\toprule
Method & Single Obj & Two Obj & Count & Color & Pos & Color Attri & Overall \\
\midrule
SDv1.5~\cite{rombach2022high}  & 0.97 & 0.38 & 0.35 & 0.76 & 0.04 & 0.06 & 0.43 \\
SDv2.1~\cite{rombach2022high} & 0.98 & 0.51 & 0.44 & 0.85 & 0.07 & 0.17 & 0.50 \\
SDXL~\cite{sdxl} & 0.98 & 0.74 & 0.39 & 0.85 & 0.15 & 0.23 & 0.55 \\
PixArt-$\alpha$~\cite{chen2023pixartalpha} & 0.98 & 0.50 & 0.44 & 0.80 & 0.08 & 0.07 & 0.48 \\
DALL-E 2~\cite{dalle2}  & 0.94 & 0.66 & 0.49 & 0.77 & 0.10 & 0.19 & 0.52 \\
DALL-E 3~\cite{dalle3}  & 0.96 & 0.87 & 0.47 & 0.83 & 0.43 & 0.45 & 0.67 \\
\midrule
Chameleon~\cite{team2024chameleon}  & - & - & - & - & - & - & 0.39 \\
LWM~\cite{liu2024world}  & 0.93 & 0.41 & 0.46 & 0.79 & 0.09 & 0.15 & 0.47 \\
SEED-X~\cite{ge2024seed}  & 0.97 & 0.58 & 0.26 & 0.80 & 0.19 & 0.14 & 0.49 \\
Show-o~\cite{xie2024show} & 0.95 & 0.52 & 0.49 & 0.82 & 0.11 & 0.28 & 0.53 \\
JanusFlow~\cite{ma2025janusflow} & 0.97 & 0.59 & 0.45 & 0.83 & \textbf{0.53} & 0.42 & 0.63 \\
\textbf{Ours} & \textbf{0.99} & \textbf{0.89} & \textbf{0.84} & \textbf{0.87} & 0.35 & \textbf{0.56} & \textbf{0.75} \\
\bottomrule
\end{tabular}
}
\end{table}

\begin{table*}[t]
\caption{\textbf{T2V performance on VBench~\cite{huang2024vbench}.} Quantitative results are quoted from the official Vbench Leaderboard.}
\label{tab:vbench_forced}
\resizebox{\textwidth}{!}{%
\begin{tabular}{lcccccccccc}
\toprule
\multicolumn{1}{c}{}                                                    & \multicolumn{3}{c}{\textbf{Overall Scores}}        & \multicolumn{5}{c}{\textbf{Technical Quality}}                                           & \multicolumn{2}{c}{\textbf{Aesthetic Quality}}             \\ \cmidrule(lr){2-4} \cmidrule(lr){5-9} \cmidrule(lr){10-11}
\textbf{Model Name}                                                     & Total Score    & Quality          & Semantic       & Subject & Background      & Temporal & Motion  & Dynamic & Aesthetic    & \multicolumn{1}{c}{Imaging} \\ \midrule
EasyAnimateV5.1~\cite{xu2025easyanimate}          & \textbf{83.42} & 85.03            & 77.01          & 98.00            & 97.41            & 99.19            & 98.02          & 57.15          & \textbf{69.48}       & \multicolumn{1}{c}{\textbf{68.61}}  \\
MiniMax-Video-01~\cite{minimax2024video}          & 83.41          & 84.85            & 77.65          & 97.51            & 97.05            & 99.10            & 99.22          & 64.91          & 63.03                & \multicolumn{1}{c}{67.17}           \\
Kling 1.6~\cite{kuaishou2025kling}                & 83.40          & 85.00            & 76.99          & 97.40            & 96.84            & 98.64            & 99.13          & 62.22          & 64.81                & \multicolumn{1}{c}{69.70}           \\
Wan2.1-T2V-1.3B~\cite{wan2025wan}                 & 83.31          & \textbf{85.23}   & 75.65          & 97.56            & \textbf{97.93}   & 99.55            & 98.52          & 65.19          & 65.46                & \multicolumn{1}{c}{67.01}           \\
HunyuanVideo~\cite{kong2024hunyuanvideo}          & 83.24          & 85.09            & 75.82          & 97.37            & 97.76            & 99.44            & 98.99          & \textbf{70.83} & 60.36                & \multicolumn{1}{c}{67.56}           \\
Gen-3~\cite{runway2024gen3}                       & 82.32          & 84.11            & 75.17          & 97.10            & 96.62            & 98.61            & \textbf{99.23} & 60.14          & 63.34                & \multicolumn{1}{c}{66.82}           \\
Vchitect-2.0 (VEnhancer)~\cite{fan2025vchitect}   & 82.24          & 83.54            & 77.06          & 96.83            & 96.66            & 98.57            & 98.98          & 63.89          & 60.41                & \multicolumn{1}{c}{65.35}           \\
CogVideoX1.5-5B~\cite{yang2024cogvideox}          & 82.17          & 82.78            & \textbf{79.76} & 96.87            & 97.35            & 98.88            & 98.31          & 50.93          & 62.79                & \multicolumn{1}{c}{65.02}           \\ \midrule
\textbf{Ours}                                                           & 83.00          & 84.27            & {77.92} & \textbf{98.39}   & 97.68            & \textbf{99.87}   & {99.10} & 56.67          & 62.48                & \multicolumn{1}{c}{64.56}           \\ \midrule
\multicolumn{1}{c}{}                                                               & \multicolumn{8}{c}{\textbf{Semantic Fidelity}}                                                                                                                 \\ \cmidrule(lr){2-10}
\textbf{Model Name}          & Object   & Multi-Obj & Action   & Color            & Spatial & Scene            & Appearance  & Temporal & Overall   &                                     \\ \midrule
EasyAnimateV5.1~\cite{xu2025easyanimate}          & 89.57          & 66.85            & 95.60          & 77.86            & 76.11            & 54.31            & 23.06          & 24.61          & 26.47                &                                     \\
MiniMax-Video-01~\cite{minimax2024video}          & {87.83} & \textbf{76.04}   & 92.40          & 90.36            & 75.50            & 50.68            & 20.06          & 25.63          & 27.10                &                                     \\
Kling 1.6~\cite{kuaishou2025kling}                & 93.34          & 63.99            & 96.20          & 81.26            & 79.08            & 55.57            & 20.75          & 24.51          & 26.04                &                                     \\
Wan2.1-T2V-1.3B~\cite{wan2025wan}                 & 88.81          & 74.83            & 94.00          & 89.20            & 73.04            & 41.96            & 21.81          & 23.13          & 25.50                &                                     \\
HunyuanVideo~\cite{kong2024hunyuanvideo}          & 86.10          & 68.55            & {94.40} & \textbf{91.60}   & 68.68            & 53.88            & 19.80          & 23.89          & 26.44                &                                     \\
Gen-3~\cite{runway2024gen3}                       & 87.81          & 53.64            & 96.40          & 80.90            & 65.09            & 54.57            & 24.31          & 24.71          & 26.69                &                                     \\
Vchitect-2.0 (VEnhancer)~\cite{fan2025vchitect}   & 86.61          & {68.84}   & \textbf{97.20} & 87.04            & {57.55}   & \textbf{56.57}   & 23.73          & {25.01} & \textbf{27.57}       &                                     \\
CogVideoX1.5-5B~\cite{yang2024cogvideox} & {87.47} & {69.65}   & \textbf{97.20} & {87.55}   & \textbf{80.25}   & {52.91}   & \textbf{24.89} & {25.19} & 27.30                &                                     \\ \midrule
\textbf{Ours}                                                           & \textbf{93.54} & 71.06            & 93.60          & {88.89}   & {73.15}   & {44.33}   & {23.45} & \textbf{25.81} & {26.99}       &  \\
\bottomrule
\end{tabular}
}
\end{table*}

\paragraph{Text-to-Image results.}
Fig.~\ref{fig:t2i} presents the qualitative results of our \ourmethod on text-to-image generation tasks.
We can observe that the generated images exhibit remarkable visual fidelity across diverse domains, including human portraits, animals, and landscapes. 
For human faces, \ourmethod produces high-resolution details such as realistic skin textures and nuanced expressions. %
In animal depictions, intricate features like fur patterns and feather details are vividly rendered. 
Additionally, landscape generations showcase natural lighting effects and coherent structural compositions without artifacts.

\begin{figure*}[htp]
  \centering
  \includegraphics[width=1\textwidth]{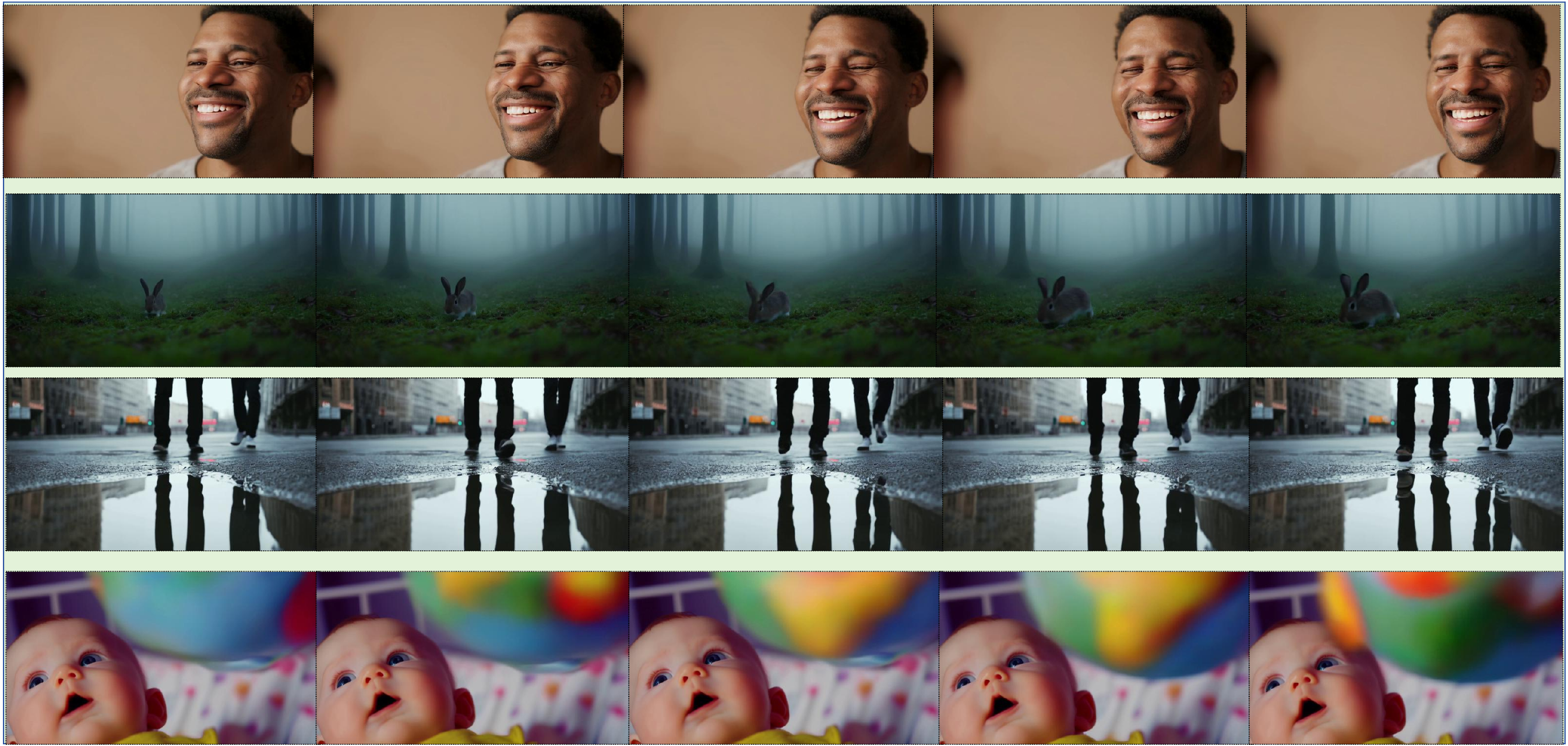}
  \caption{
  \textbf{Example video sequences generated by our proposed \ourmethod on text-to-video settings.}
  }
  \label{fig:t2v}
\end{figure*}

\paragraph{Text-to-Video results.}
Fig.~\ref{fig:t2v} shows the qualitative results of our \ourmethod on text-to-video generation tasks, indicating our model's ability to produce temporally coherent and dynamically rich outputs.
For instance, samples from the first-row samples demonstrate fine-grained facial performance: micro-expressions like orbicularis oculi contractions during laughter are captured with anatomical accuracy, while the camera's lateral dolly motion maintains subject framing without introducing judder.%
Moreover, second-row footage achieves animal-specific locomotion realism—the rabbit exhibits proper quadrupedal gallop dynamics (suspensory phase clarity) alongside environment-reactive fur movement, with fog interaction showing consistent volumetric lighting transitions. \
Additionally, third-row sequences highlight complex urban reflections: synchronized footstep splashes create plausible ripples in the puddle, and the low-angle tracking shot preserves silhouette continuity between pedestrians despite varying stride lengths.

\begin{figure*}[htp]
  \centering
  \includegraphics[width=1\textwidth]{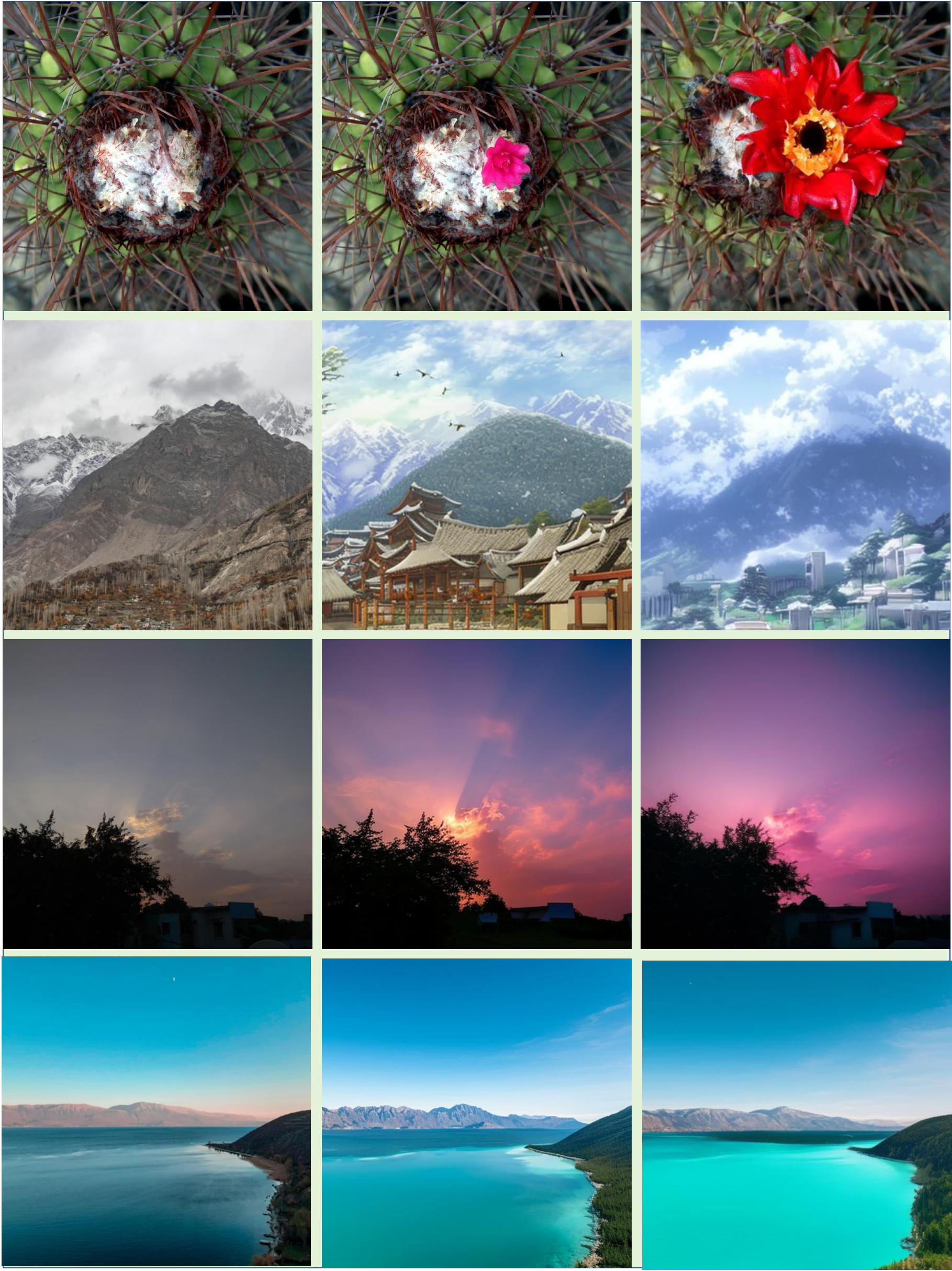}
  \caption{
  \textbf{Example edited images generated by our proposed \ourmethod on image-to-image editing settings.}
  For each row, the leftmost represents the source image provided, and the middle column is the ground truth, the rightmost image is the edited output from our model.
  }
  \label{fig:i2i}
\end{figure*}

\paragraph{Image-to-Image editing.}
In addition to traditional text-to-image/video tasks, we further extent our \ourmethod to support detail-preserving image and video editing tasks.
The qualitative results of image-to-image editing are illustrated in Fig.~\ref{fig:i2i}.
For each row, the leftmost represents the source image provided, and the middle column are the ground truth, the rightmost images are edited output from our model.
As shown in the first row, our method successfully introduces dynamic botanical elements while meticulously preserving structural integrity—notice the cactus spine details remain sharp even as a hyper-realistic bloom emerges from the center. 
The second-row samples demonstrate robust cross-domain style translation: the original mountain  landscape retains its topographical accuracy even after transitions to anime aesthetics and futuristic environments, with snow textures maintaining consistency across stylistic shifts.

Furthermore, in the sky editing sequences (third row), our system achieves seamless temporal coherence in atmospheric transitions—cloud formations maintain continuity from overcast to dusk states, while the horizon line stays geometrically consistent as hues shift between magenta and indigo gradients.
Finally, the lake edits (bottom row) exhibit precise perspective-aware expansion: the shorelines extend naturally under varying angles, and the water surface tension adapts authentically to different lighting conditions (midday azimuth vs. twilight reflections), without introducing tessellation artifacts.

\begin{figure*}[htp]
  \centering
  \includegraphics[width=1\textwidth]{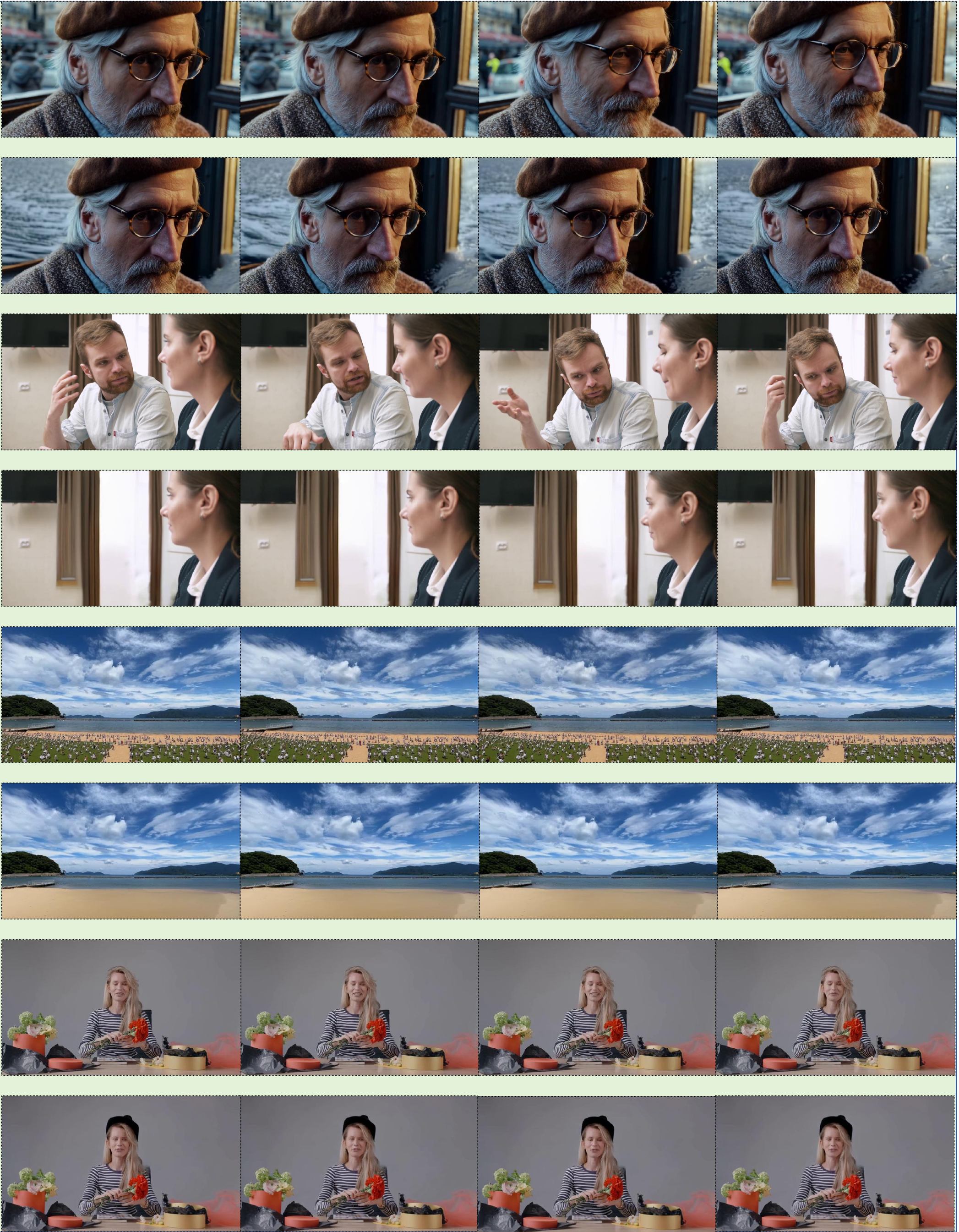}
  \caption{
  \textbf{Example edited videos generated by our proposed \ourmethod on video-to-video editing settings.}
  For each pair, the upper one represents the source videos provided by users, and and bottom videos are edited output from our model.
  }
  \label{fig:v2v}
\end{figure*}

\paragraph{Video-to-Video editing.}
Regarding video-to-video editing, the qualitative results presented in Fig.~\ref{fig:v2v} demonstrate \ourmethod's versatile adaptability across diverse spatiotemporal manipulation tasks.
Obviously, our framework uniformly supports object removal, background replacement, attribute addition, \emph{etc,}  while maintaining cross-frame consistency.
For instance, the background of the old man is changed from street to an ocean, the man besides the woman (people and grasses on the beach) is removed, and a hat is added to the woman in Fig.~\ref{fig:v2v}.
Crucially, these compound edits preserve microscopic details with no flickering in edited regions and wind-responsive elements, achieving a cross-frame consistency score that underscores the method's unified adaptability for complex multi-operation editing scenarios.

\begin{figure*}[htp]
  \centering
  \includegraphics[width=1\textwidth]{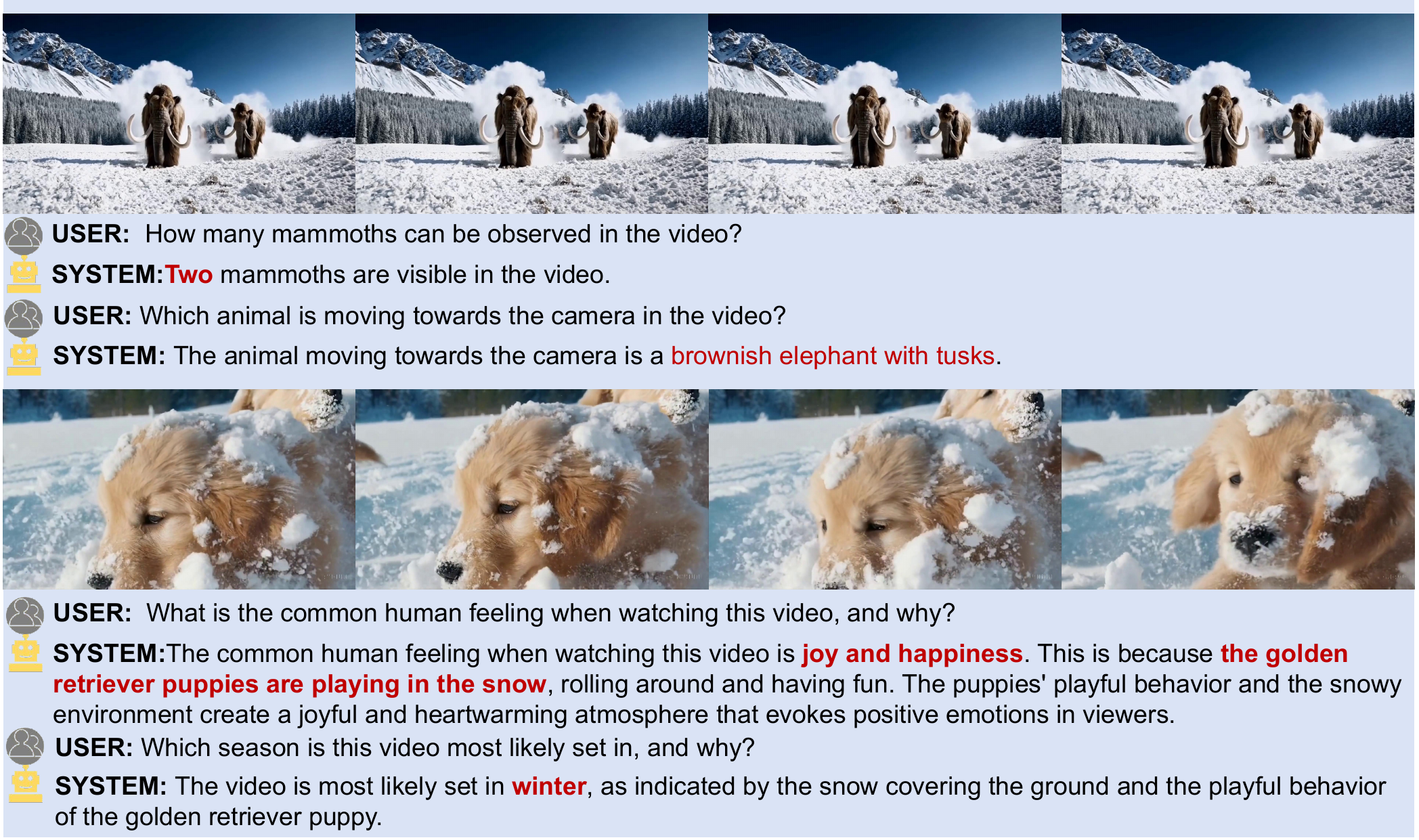}
  \caption{
  \textbf{Example responses generated by our proposed \ourmethod on video understanding settings.}
  }
  \label{fig:understanding}
\end{figure*}

\paragraph{Video understanding.}
Furthermore, we evaluate the video understanding performance of our \ourmethod in Fig.~\ref{fig:understanding}, where the system demonstrates robust multimodal reasoning across diverse spatiotemporal contexts. 
In the Pleistocene snowscape sequence, it accurately grounds quantitative analysis ("two mammoths") while distinguishing species-specific traits ("brownish elephant with tusks") and inferring motion dynamics (animal approaching the camera). 
For the companion winter scene featuring golden retriever puppies, the method identifies seasonal context ("winter, as indicated by snow cover") and derives emotional valence ("joy and happiness") from behavioral cues (playful rolling, social interaction), even when occlusions occur during rapid canine movements. 
These results underscore \ourmethod's ability to simultaneously process object attributes, temporal action patterns, and environmental metadata—key competencies for unified video grounding and interpretation. 
The consistent performance across paleontological and domestic contexts highlights its domain-agnostic adaptability.

\begin{figure*}[htp]
  \centering
  \includegraphics[width=1\textwidth]{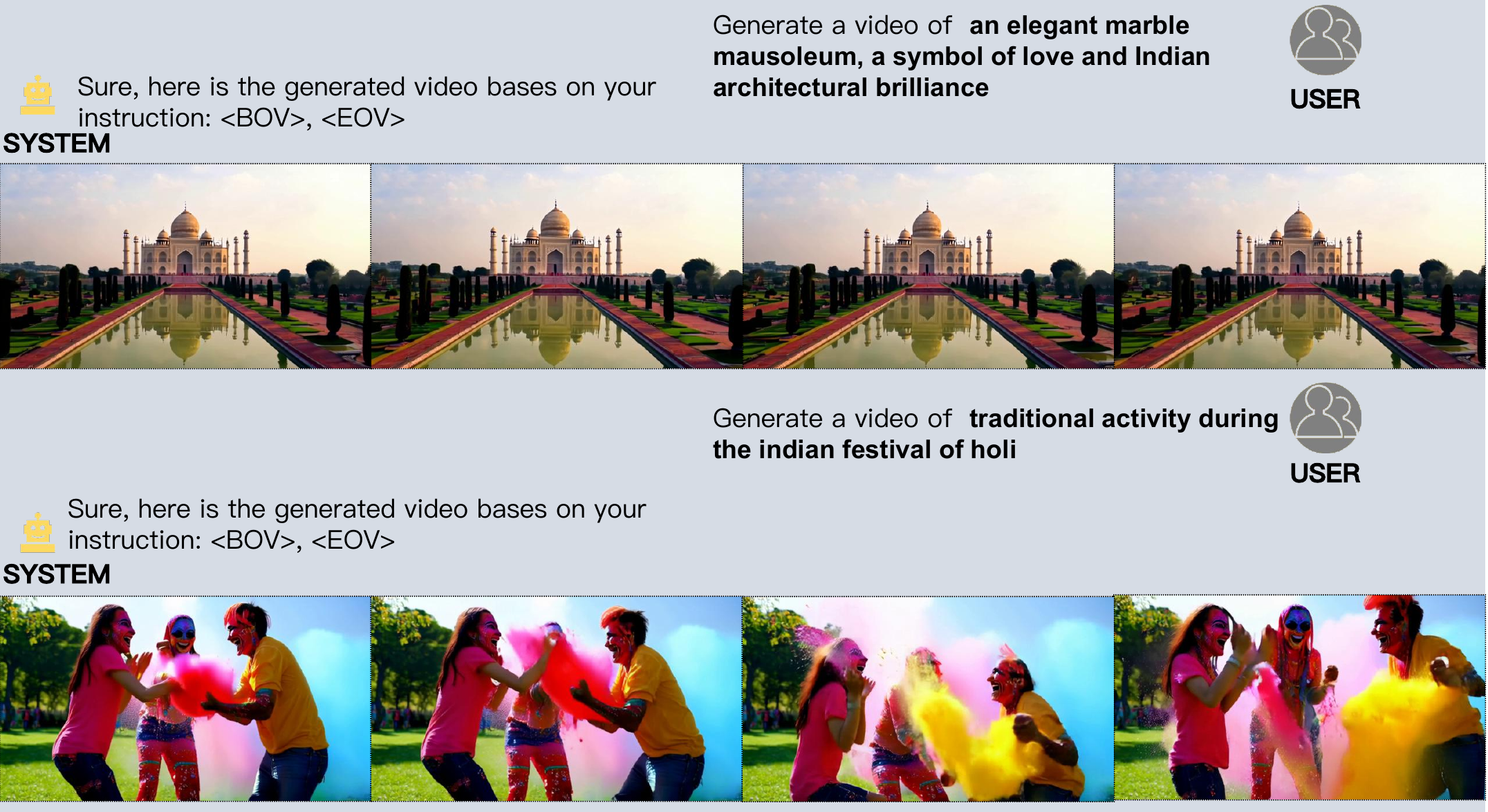}
  \caption{
  \textbf{Example videos generated by our proposed \ourmethod under \emph{think mode} for text-to-video tasks that require thinking with user instructions.}
  }
  \label{fig:think}
\end{figure*}

\paragraph{Think mode.}
Finally, Fig.~\ref{fig:think} present the videos produced by our system under ``think" mode, in which the model is asked to ``think" with a long chain-of-thought to fully understand user instructions and produce corresponding visual output.
For example, when generating the mausoleum sequence, our model first parses ambiguous descriptors—interpreting ``elegant marble" as requiring stratified erosion textures and ``Indian architectural brilliance" as necessitating precise dome geometry—before translating these inferences into optimized visual tokens. 
This manifests as calibrated material properties (light refraction indices in marble veins), architecturally accurate proportions (minaret height-to-width ratios), and environmentally responsive details (wind-aligned water ripples). 
\section{Limitations}
Despite successfully unifying video understanding, generation and editing in a single framework via connecting MLLMs with diffusion decoder, our approach faces persistent limitations rooted in data scalability, computational constraints, and architectural fragmentation.
The current implementation struggles with high-fidelity, long-duration video synthesis due to prohibitive memory demands and limited training data for complex interactions and distribution modeling.
More fundamentally, the decoupled design between diffusion decoders and MLLMs creates optimization misalignments of latent representations optimized for generative tasks fail to seamlessly transfer to discriminative understanding tasks, particularly in video domains requiring frame-by-frame causal reasoning.
We plan to continuously address these limitations through systematic enhancements to our framework. 
Ongoing efforts might include refining the diffusion-MLLM interface via shared optimization, developing adaptive training protocols for complex physical interactions, and scalable data curation pipelines to gather higher-quality training data.
%

%


\bibliographystyle{unsrt}
\bibliography{ref.bib}

\newpage
\appendix
\section{Prompts Used in the Main Text}
Here we list all the prompts used in the main text. 
The prompts are arranged in the order of their appearance in the figure, from top to bottom and from left to right.
\begin{tcolorbox}
\begin{center}
    \small
    \textbf{Prompts used in Figure 1}
\end{center}
\tiny
\textbf{Text-to-Image:}
\begin{itemize}
    \item p1: A volcanic springs steaming in Icelandic winters. Azure pools steam among obsidian rocks where snowflakes vanish near bubbling geothermal vents.
    \item p2: A serene scene of a misty morning in a valley, with a calm river flowing through it. The sunlight is shining through the trees, casting a warm glow on the landscape. The river is surrounded by lush green trees and a forest, creating a peaceful atmosphere.
    \item p3: An image features Finnish icebreaker ships plowing Arctic channels. Steel prows split blue-white plates forming frozen wakes under polar twilight streaked with sun dogs.
    \item p4: A black and white photograph of a serene mountain lake surrounded by trees. The mountain can be seen in the distance, towering over the lake. The reflection of the mountain and its snowy peak can be seen clearly in the calm waters of the lake.
    \item p5: Antarctic icebergs sculpted by polar winds. Azure caverns glow in monumental ice cliffs floating through silver seas where penguins porpoise through waves.
    \item p6: A futuristic image featuring a woman with glowing yellow eyes. She is wearing a black and gold outfit, which is adorned with intricate patterns and lights. The woman has long hair, and her eyes are prominently glowing yellow. She is the main focus of the scene.
    \item p7: An image that captures the essence of retro chic. At the center of the frame is a woman exuding a sense of confidence and style.
    \item p8: An image depicts a dystopian, cyberpunk-inspired scene. In the foreground, a figure wearing an orange jacket with a blue and black helmet is standing, facing the viewer. The figure's helmet is adorned with a blue circular element. The buildings exhibit a dilapidated appearance, with visible signs of wear and rust.
    \item p9: An image captures a woman wearing sunglasses and a jacket, standing on a boat and smiling. She is wearing a backpack, which is located on her back. The boat is floating on a river, and the water is visible in the background.
    \item p10: An image features a golden retriever playing in autumn leaves, tongue lolling. Crisp red and orange foliage blankets the ground, with a weathered wooden fence and distant misty mountains framing the scene.
\end{itemize}

\textbf{Text-to-Video:}
\begin{itemize}
    \item p1: A close-up shot of a black pan on a stovetop, where two round, golden-brown falafel balls are being fried. The falafel balls are speckled with green herbs and appear crispy on the outside. As the video progresses, a hand enters the frame and flips one of the falafel balls, revealing a similarly cooked side. The hand then removes the fried falafel from the pan, leaving the other ball to continue frying.
    \item p2: The process of pouring a liquid into a metal cup. The liquid appears to be a dark brown color, possibly a type of coffee or tea. The cup is cylindrical with a flared top and a black handle. The liquid is poured from a clear glass container with a handle. The background is blurred, but it seems to be an indoor setting with a dark surface. The style of the video is a close-up shot, focusing on the action of pouring the liquid into the cup.
    \item p3: A person is seen using a power tool to grind a piece of metal. The person is wearing black gloves and is holding the power tool with their right hand. The metal is being held in place by a blue vise. As the person grinds the metal, sparks fly out from the point of contact. The background is blurred, but it appears to be a workshop environment.
\end{itemize}

\textbf{Image-to-Image:}
\begin{itemize}
    \item p1: redesign this picture into tourism illustration
    \item p2: Wipe off the bird from the photo.
\end{itemize}

\textbf{Video-to-Video:}
\begin{itemize}
    \item p1: Add a hat.
\end{itemize}
\end{tcolorbox}

\begin{tcolorbox}
\begin{center}
    \small
    \textbf{Prompts used in Figure 5}
\end{center}
\tiny

p1: Finnish forest reindeer foraging lichen in snowbound woods. Antlers crown among pine shadows where snowshoe tracks circle frozen lakes. \\
p2: Clouds shimmering over fjord cliffs. Electric-blue wisps ladder midnight skies where satellites streak above inky waters holding mirror galaxies. \\
p3: Icelandic glacier caves radiating sapphire light. Frozen corridors wind through ancient ice, shimmering turquoise under headlamps near subterranean waterfalls. \\
p4: A biplane soaring above cotton-cumulus clouds. Ragtime echoes from propeller buzz over patchwork landscapes where toy windmills spin on distant hillsides. \\
p5: A beautiful girl with flowing chestnut hair and emerald eyes, wearing a sunflower-print sundress. She stands in a lavender field at sunset, holding a vintage book, with golden light highlighting honeybees hovering near purple blooms. \\
p6: A large, well-lit bedroom with a bed positioned towards the right side of the room. The bed is adorned with a gold comforter, giving it a luxurious appearance. The room also features a chair on the left side and another chair closer to the center of the room. The overall atmosphere of the bedroom is elegant and inviting. \\
p7: A small black and white dog standing on a sandy beach. The dog appears to be looking to the side, possibly observing something or someone in the distance. The dog is wearing a collar, which adds a sense of charm to the scene.  \\
p8: A large elephant standing in a field with trees and bushes. The elephant is positioned near the center of the scene, surrounded by a variety of trees and bushes. Some of the trees have no leaves, while others have a mix of green and brown leaves. The elephant appears to be grazing or exploring the area, possibly searching for food. \\
p9: A ginger cat perched on a moss-covered stone wall, amber eyes wide. Dappled sunlight filters through oak leaves onto its striped fur, with bluebells and ferns crowding the wall’s base. \\
\end{tcolorbox}

\begin{tcolorbox}
\begin{center}
    \small
    \textbf{Prompts used in Figure 6}
\end{center}
\tiny
p1: A video shows a man's face breaking into a warm, genuine smile as he recognizes someone off-camera. The style is candid and heartwarming. The lighting is soft and natural. The overall impression is one of sudden, happy recognition. \\
p2:  A rabbit hopping through a misty forest floor at dawn, its movements quiet and cautious. The style is atmospheric and mysterious, with the rabbit as a gentle guide through the scene. The lighting is the soft, diffused light of a foggy morning. The overall impression is one of magical, quiet exploration. \\
p3: A man's reflection in a puddle on a city street as he walks by. The style is artistic and anonymous. The lighting is the reflected light of the city. The overall impression is one of a fleeting moment in the urban flow. \\
p4: A video shows a baby's wide, fascinated eyes as he look up at a colorful mobile spinning above their crib. The style is wondrous and focused, capturing their developing senses. The lighting is soft and colorful, perhaps from the mobile itself. The overall impression is one of a whole new world opening up. \\
\end{tcolorbox}

\begin{tcolorbox}
\begin{center}
    \small
    \textbf{Prompts used in Figure 7}
\end{center}
\tiny
p1: Add a red flower to the center of the cactus plant. \\
p2: Change it into kyoto animation \\
p3: Change the color of the clouds to pink \\
p4: Change the color of the water to turquoise \\
\end{tcolorbox}

\begin{tcolorbox}
\begin{center}
    \small
    \textbf{Prompts used in Figure 8}
\end{center}
\tiny
p1: Add a sea.\\
p2: Eliminate the man. \\
p3: Change the park to beach. \\
p4: Add a hat to the woman. \\
\end{tcolorbox}

\end{document}